\pdfoutput=1

\documentclass[11pt]{article}

\usepackage[]{acl}

\usepackage{times}
\usepackage{adjustbox}
\usepackage{multirow}
\usepackage{latexsym}
\usepackage{amsmath}
\usepackage{amssymb}
\usepackage{makecell}
\usepackage{amssymb}

\usepackage{mathabx}
\usepackage[T1]{fontenc}

\usepackage[utf8]{inputenc}

\usepackage{listings}
\usepackage{xcolor}

\usepackage{microtype}

\title{\textbf{MateICL}: Mitigating Attention Dispersion in Large-Scale In-Context Learning}

\author{Murtadha Ahmed,  Wenbo, Liu yunfeng \\
        Zhuiyi AI Lab, Shenzhen, China}

\begin{document}
\maketitle

\begin{abstract}
Large Language Models (LLMs) have demonstrated remarkable capabilities in In-Context Learning (ICL). However, the fixed position length constraints in pre-trained models limit the number of demonstration examples. Recent efforts to extend context suffer from attention dispersion as the number of demonstrations increases.
In this paper, we introduce Mitigating Attention Dispersion in large-scale ICL (MateICL) that enables LLMs to maintain effective self-attention as the context size grows. We first split the context into multiple windows, each filled to the model's context capacity, which are processed separately. Then, we introduce an additional layer to recalibrate the attention weights, prioritizing the query tokens as the number of demonstrations increases. Our empirical results show that MateICL can effectively leverage larger contexts to improve ICL performance. Compared to retrieval-based baselines, MateICL consistently achieves better performance without requiring an externally trained retrieval model. Despite recent advances in inference strategies (e.g., 32k token contexts), our results demonstrate that MateICL remains beneficial in computationally resource-constrained settings. The code is publicly available at 
\href{https://github.com/amurtadha/MateICL}{MateICL}

\end{abstract}


\section{Introduction} \label{sec:introduction}

Large Language Models (LLMs) have shown notable In-Context Learning (ICL) abilities. ICL is a paradigm that enables LLMs to adeptly perform a variety of novel tasks based on contextual demonstration examples without the need for fine-tuning step. This approach eliminates the need for parameter updates within the models \cite{BrownMRSKDNSSAA20,HanZDGLHQYZZHHJ21,Qiu-2003-08271}. As a result, it not only boosts computational efficiency but also represents a major step towards achieving general artificial intelligence. The relevance and adoption of ICL have surged in line with the continuous expansion and increasing complexity of LLMs \cite{Zhang-2205-01068,Chowdhery-2204-02311,abs-2302-13971}.
\begin{figure}[t]
	\centering
	\includegraphics[scale=0.7]{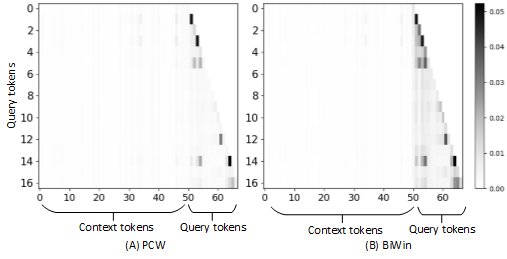}
	\caption{Visualization of attention weights for a context window (\(\text{W}=9\)) on the TREC Dataset using LLaMA-7B, comparing PCW (A) \cite{pcw} and MateICL (B), where tokens \(> 50\) represent the query. 
	}
	\label{fig:att}
	
\end{figure}
Traditional ICL methods   restricts the number of demonstration examples to fill within the context window size of the specific LLM (e.g., 1024 tokens in GPT2) \cite{ZhaoWFK021}. 
To address this limitation, \citet{Garg0001TLV22,MinLZH22,ChenZZK022} improve ICL through meta-learning and fine-tuning on downstream tasks, but the limited diversity of annotated tasks and biases hinder generalization. Alternatively, retrieval-based methods focus on independently training task-specific retrievals, which are then used to select the relevant context  \cite{LiLYLZNXWQ23,llm-r}. While they show promising results,  the need to train an external retrieval  for each task significantly limit their practical application in real-world scenarios \cite{icl_survey_2024}.
Another line of research has explored various approaches to retraining long-range language models with extrapolation, extending them to 128 times the limit of existing  LLMs \cite{Mukai-2302-04931,Gu0WH23}. However, these approaches require additional training over several steps, which can be time-consuming.

The Parallel Context Window (PCW) \cite{pcw} enhances ICL by distributing demonstration examples across multiple context windows \(\text{W}\), each filled with numerous examples within the LLM’s context capacity, which are then individually encoded. 
Unlike context averaging \cite{YangLMZDT24} or context voting \cite{nbce}, PCW requires the query (i.e., the test example) to attend to the entire context, ensuring comprehensive knowledge transfer. However, as shown in Fig.\ref{fig:att} and theoretically explained in Sec.\ref{sec:attention}, an increase in the number of demonstrations reduces the attention weights of the query tokens. This attenuation limits the scalability of the context to a certain number of windows, with a notable threshold at \(\text{W} > 3\).

In this paper, we introduce Mitigating Attention Dispersion in large-scale ICL (MateICL)  that allows LLMs to maintain effective self-attention as the number of demonstrations increases.
We introduce an additional layer that recalibrates the attention weights to prioritize the query tokens. Note that parallel context leads to a linear increase in encoding complexity relative to the number of windows, as opposed to the quadratic complexity encountered when processing all examples simultaneously. 
\textbf{Despite recent advances in inference strategies (e.g., 32k context tokens), MateICL remains practically valuable for two reasons:} (1) It mitigates attention dispersion by splitting the context into separate windows, which are inferred independently and then integrated, preventing distraction from too many examples. (2) It is particularly useful in resource-constrained environments where handling large contexts is challenging.

In brief, our main contributions are as follows:
\begin{enumerate}
\item We propose the Mitigating Attention Dispersion in large-scale ICL (MateICL) framework, which enables models to maintain effective self-attention as the number of demonstrations increases. Specifically, we introduce an additional layer that recalibrates attention weights, encouraging the model to focus more on the generated tokens in large-context settings.

\item We conduct extensive experiments across various benchmark NLP tasks, demonstrating that MateICL effectively scales the number of demonstrations while improving model stability and performance.
\end{enumerate}

\section{Related Work}\label{sec:related_work}
%

ICL has emerged as a prominent focus of research, first introduced by \cite{BrownMRSKDNSSAA20}. \cite{ZhaoWFK021} observed the instability in LLM predictions, where the models often favor certain outputs disproportionately. Subsequent works by \cite{LuBM0S22} and \cite{HanH0SW23} showed that performance is sensitive to various factors, including the prompt format, the set of training examples, and their order. To mitigate these challenges, \cite{ZhaoWFK021} proposed a method for estimating model biases and adjusting calibration parameters to balance output likelihoods. 
The authors of \cite{MinLHZ22} explored decision boundary adjustments via prototypical cluster distribution. The authors of \cite{HazarikaNH22} introduced techniques for controlling small Encoder-Decoder models in zero-shot settings using attention biasing and context augmentation. Advances in prompt engineering have optimized demonstration order \cite{LuBM0S22}, enhancing the diversity.

\subsection{Demonstration Retrieval}

{Demonstration retrieval} is an information retrieval technique that utilizes dense vectors for semantic matching between queries and documents in the latent space \cite{wang2022simlm}. Unlike sparse methods such as BM25, dense retrieval capitalizes on the capabilities of LM  to overcome vocabulary mismatches \cite{bert}. Further, hard negative mining \cite{karpukhin2020dense}, knowledge distillation \cite{ren2021rocketqa}, and continual pre-training \cite{wang2022simlm} have further enhanced its performance. \textit{Retrieval-Augmented LLMs} combine the generative power of LLMs with the ability to retrieve relevant external information,
including input concatenation \cite{shi2023replug}, intermediate attention fusion \cite{borgeaud2022improving}, and output interpolation \cite{khandelwal2020generalization}. These retrieval-augmented methods aim to boost LLM performance on downstream tasks by retrieving and incorporating informative examples \cite{LiLYLZNXWQ23, luo2023context}. For instance, \cite{Ye0F0K23} introduced CEIL, which uses determinantal point processes and a contrastive learning objective to model interactions between inputs and examples, guiding model preferences. LLM-R \cite{llm-r} proposed an iterative framework for training a reward model based on LLM feedback to evaluate the quality of candidate examples. However, these retrieval-based approaches often require additional task-specific training, which can be both time-consuming and resource-intensive. In contrast, we propose scaling the ICL context to ensure diversity, which enhances performance stability.

\subsection{Context Extension}

Efforts to expand the contextual understanding of LLMs have led to two primary strategies: fine-tuning and few-shot learning. Sparse attention methods \cite{ZaheerGDAAOPRWY20, GuoAUONSY22} have been developed to address the memory limitations in handling longer inputs. The authors of \cite{PressSL22} introduced relative positional information to enhance extrapolation, albeit at the cost of increased computational overhead. The SLED method \cite{abs-2208-00748} offers an efficient solution for processing long texts by encoding overlapping segments, similar to the Fusion-in-Decoder approach \cite{IzacardG21}, though its effectiveness requires additional training, which can be time-intensive. To avoid fine-tuning, \cite{Yaru-2212-06713} proposed structured prompting, where demonstration examples are jointly attended to by the test example via a scaled attention mechanism. PCW \cite{pcw} bypassed the need for scaled attention by introducing specially designed positional indexing mechanisms that treat demonstration examples individually. However, the performance of these methods tends to degrade as the number of demonstration examples increases, leading to a decline in the quality of generated outputs. To address this, we propose recalibrating the attention weights to prioritize the query tokens as the number of demonstration examples increases.

\section{MateICL}\label{sec:approach}	
	\begin{figure*}
	\centering
	\includegraphics[scale=0.7]{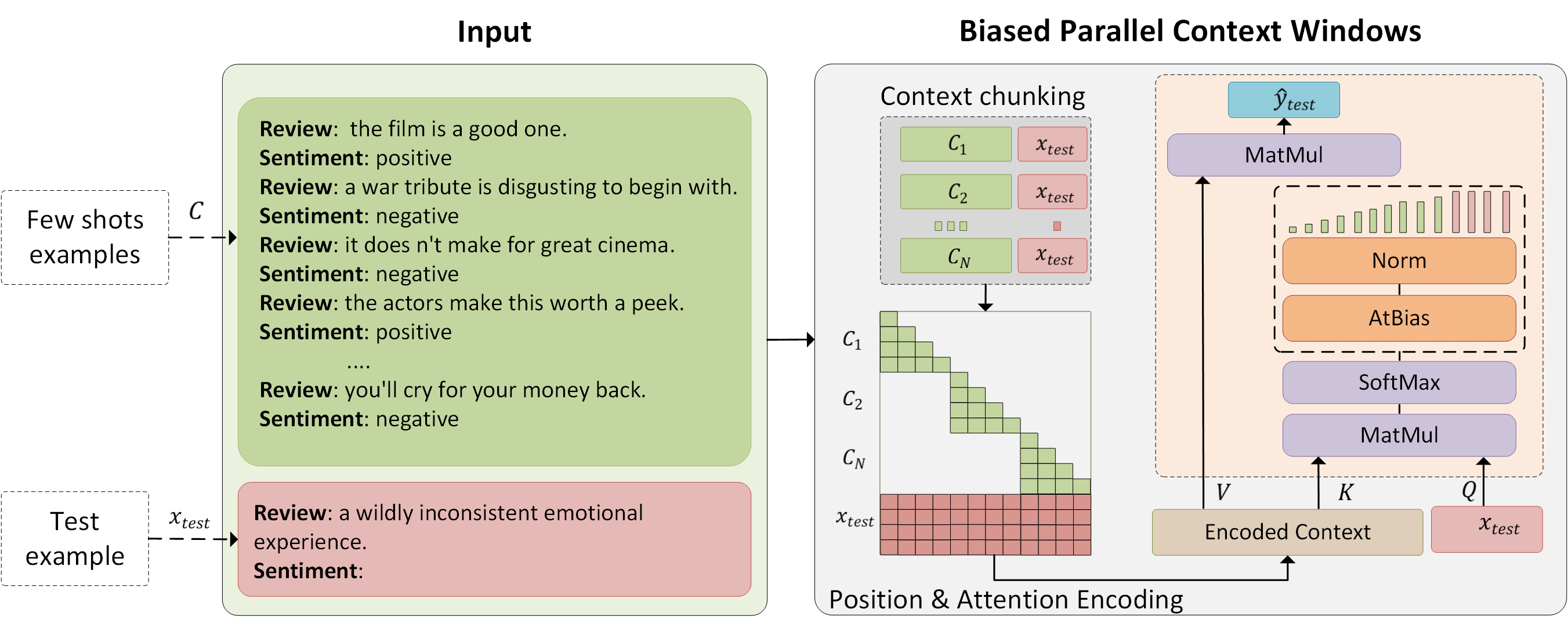}
	
	\caption{An example of our MateICL.  The context is divided into multiple windows to match the LLM's capacity (e.g., up to 1024 tokens for GPT-2). Each window is independently encoded, restricting windows to only attend to their own tokens. MateICL adds an additional ``AtBias'' layer to ensure that the task tokens can still attend to themselves as the number of examples grows.}
	\label{fig:bpcw}
\end{figure*}	
In this section, we describe our MateICL, shown in Fig.\ref{fig:bpcw}.  Specifically, we first present parallel context  mechanism, which enables the encoding of large contexts. We then describe our approach to maintaining the model’s self-attentiveness as the context expands.

\subsection{Parallel Context Windows}
In ICL, we categorize the input tokens into context tokens \((C)\) and task tokens \((T)\). The context tokens, \(C\), consist of demonstration examples that assist the model \(\mathcal{M}\) in performing tasks, while task tokens, \(T\), represent the query or test examples. In standard ICL, the context size is defined as \((C = N - T)\), where \(N\) represents the maximum context capacity of \(\mathcal{M}\). To enable more demonstration,
PCW \cite{pcw} introduces to divide \(C\) into multiple windows \((W)\), each containing up to \(k\) sentences, which fall within the capacity of \(\mathcal{M}\), depicted on the left side of Fig.\ref{fig:bpcw}.  PCW allows extending up to  \((W \times C + T)\) tokens by assigning one of \(N\) to each position \(i\) in the sequence  as follows:
\begin{equation}
		\mathbf{p}^{PCW}_i = 
		\adjustbox{width=0.8\linewidth}{%
			$\begin{cases}
				\mathbf{p}_{(i-1 \bmod  N)+1}, & \text{if } 1 \leq i \leq WC \\
				\mathbf{p}_{i- (W-1)C}, & \text{if } WC < i \leq WC + T,
			\end{cases}$
		}
\end{equation}
where  \(\mathbf{p}(.)\) denotes the position index and \(\bmod\) for the modulo operation. Each \(C_i\) can  only attend to the tokens within the same window:
\begin{equation}
		a_{i,i'}^{w,w'} =
		\adjustbox{width=0.8\linewidth}{%
			$\begin{cases}
				1, & \text{if }  1 \leq i' \leq i \leq C \text{ and } w = w' \\
				0, & \text{otherwise}.
			\end{cases}$}
	\end{equation}
	The task tokens \(T\) must attend to \textbf{ the entire context \(C\)}:
	\begin{equation}
		a_{i,i'}^{w,w'} =
		\adjustbox{width=0.8\linewidth}{%
			$
			\begin{cases}
				1, & \text{if } 1 \leq i' \leq i \leq N \text{ and } w' \in [W+1] \\
				0, & \text{otherwise}.
			\end{cases}$
		}
		\label{eq:att_task}
	\end{equation}

According to Eq.\ref{eq:att_task}, tokens of \(T\) are required to attend to all tokens of \(C\). In the next section, we will show that as the size of \(C\) increases, the attention scores assigned to \(T\) decrease, causing the model to struggle in identifying the query. In the next section, we will show that as the size of \(C\) increases, the attention scores assigned to \(T\) decrease, causing the model to struggle in identifying the query.

\subsection{Mitigating Attention Dispersion} \label{sec:attention}
In this section, we first explain how attention becomes diluted as the number of demonstrations increases in ICL, and then present our solution to address this issue. The authors fo \cite{YuanFLWZTPWHL24} provide a theoretical explanation for this phenomenon, which we summarize below. 
For each attention head of the model $\mathcal{M}$, let $\mathbf{x}_i \in \mathbb{R}^{d_{\text{in}}}$ represent the $i$-th input token, with $U_q$, $U_k$, and $U_v$ being the projection matrices used to compute the queries, keys, and values. 
We denote $\mathbf{x}_{i\in\mathcal{D}_\text{demonstrations}} U_k$, $\mathbf{x}_{i\in\mathcal{D}_\text{demonstrations}} U_v$, $\mathbf{x}_{i\in\text{query}} U_k$, $\mathbf{x}_{i\in\text{query}} U_v$ as $\mathbf{K}_d$, $\mathbf{V}_d$, $\mathbf{K}_q$, and $\mathbf{V}_q$, respectively. To compute the response $\hat{\mathbf{y}}_r$ of each head, we have the following formulation:
	\begin{equation}\label{eq:newicl}
		\begin{aligned}
			\hat{\mathbf{y}}_r &= \text{Att}(\mathbf{x}_r U_q, \mathbf{K}_d \oplus \mathbf{K}_q, \mathbf{V}_d \oplus \mathbf{V}_q) \\ 
			&\approx \mathcal{L}\text{Att}(\mathbf{x}_r U_q, \mathbf{K}_d \oplus \mathbf{K}_q, \mathbf{V}_d \oplus \mathbf{V}_q) \\ 
			&= \mathbf{x}_r U_q \, [\mathbf{K}_d \oplus \mathbf{K}_q]^T \left[\begin{array}{c}\mathbf{V}_d \\ \mathbf{V}_q\end{array}\right]  \\ 
			&= \mathbf{x}_r U_q \mathbf{V}_q \mathbf{K}_q^T + \mathbf{x}_r U_q \mathbf{V}_d \mathbf{K}_d^T \\ 
			&= \mathbf{x}_r U_{\text{ZSL}} + \mathbf{x}_r \Delta U_{\text{ICL}}
		\end{aligned}
	\end{equation}
	where  $\mathbf{x}_r U_q \mathbf{V}_q \mathbf{K}_q^T$ represents the zero-shot learning (ZSL) output, and $\mathbf{x}_r U_q \mathbf{V}_d \mathbf{K}_d^T$ contributes to the effect of the demonstrations, which are captured by $\mathbf{x}_r U_{\text{ZSL}}$ and $\mathbf{x}_r \Delta U_{\text{ICL}}$, respectively. \(\mathcal{L}\text{Att}\) and \(\oplus\) denote linear attention an the concatenation, respectively.  Note that we approximate the standard attention with  linear attention by eliminating the softmax  for clarity \cite{dai-etal-2023-gpt}. We can rewrite this as:
	\begin{equation}
		\hat{\mathbf{y}}_r \approx \adjustbox{width=0.77\linewidth}{%
			$\underbrace{\mathcal{L}\text{Att}\left(\mathbf{x}_r U_q, \mathbf{K}_q, \mathbf{V}_q\right)}_{\text{outcome from queries}} + \underbrace{\mathcal{L}\text{Att}\left(\mathbf{x}_r U_q, \mathbf{K}_d, \mathbf{V}_d\right)}_{\text{outcome from demonstrations}}$}
	\end{equation}
This suggests that the presence of demonstrations does not directly affect the output from the queries. However, if we drop the linear approximation and revert to standard attention, we can then derive:
	\begin{equation}\label{eq:newproof}
		\begin{aligned}
			\hat{\mathbf{y}}_r &= \mathrm{Att}(\mathbf{x}_r U_q, \mathbf{K}_d \oplus \mathbf{K}_q, \mathbf{V}_d \oplus \mathbf{V}_q) \\
			&= \text{softmax}(\mathbf{x}_r U_q, [\mathbf{K}_d \oplus \mathbf{K}_q]^T) \left[\begin{array}{c} \mathbf{V}_d \\ \mathbf{V}_q \end{array}\right] \\
			&= (1-\nu(\mathbf{x}_r)) \, \text{softmax}(\mathbf{x}_r U_q \mathbf{K}_q^T) \mathbf{V}_q \\
			&\quad + \nu(\mathbf{x}_r) \, \text{softmax}(\mathbf{x}_r U_q \mathbf{K}_d^T) \mathbf{V}_d \\
			&= (1-\nu(\mathbf{x}_r)) \underbrace{\mathrm{Att}(\mathbf{x}_r U_q, \mathbf{K}_q, \mathbf{V}_q)}_{\text{outcome from queries}} \\
			&\quad + \nu(\mathbf{x}_r) \underbrace{\mathrm{Att}(\mathbf{x}_r U_q, \mathbf{K}_d, \mathbf{V}_d)}_{\text{outcome from demonstrations}}
		\end{aligned}
	\end{equation}

	
	where:
	\begin{equation}
		\nu(\mathbf{x}_r) = \adjustbox{width=0.8\linewidth}{%
			$\frac{\sum_{i} \exp\left(\mathbf{x}_r U_q \mathbf{K}_d^T\right)_i}{\sum_{i} \exp\left(\mathbf{x}_r U_q \mathbf{K}_d^T\right)_i + \sum_{j} \exp\left(\mathbf{x}_r U_q \mathbf{K}_q^T\right)_j}$}
	\end{equation}
	\textbf{As seen in Eq. \ref{eq:newproof}, $\nu(\mathbf{x}_r)$ increases with the number of demonstrations, causing the model to focus less on the query components.} This suggests that ICL may not strictly follow the scaling laws and does not equate to finetuning. Therefore, we hypothesize that an increase in the number of demonstrations may divert the model’s attention from the query, potentially leading to a performance decline.
To this end, we introduce an additional attention layer (AtBias), illustrated on the right in Fig. \ref{fig:bpcw}, which recalibrates the attention mechanism to prioritize \(T\) tokens over \(C\) tokens. Formally, to predict the response for a given query, each attention layer is designed to emphasize \(T\) tokens, ensuring their priority in the attention computation, as follows:

\begin{equation}
			\begin{aligned}
	 &\hat{\mathbf{y}}_r = \mathcal{N} (b h_r) [\mathbf{V}_d \oplus \mathbf{V}_q]\\
	 & h_r= \text{softmax}(\mathbf{x}_r U_q [\mathbf{K}_d \oplus \mathbf{K}_q]^T )
	 	\end{aligned}
	\end{equation}
	where \(\mathcal{N} (.)\) denotes the normalization function and the \(b\) is defined by:
	
	\begin{equation}
		b = 
			\begin{cases}
				\left\lfloor \frac{\text{W}}{3} \right\rfloor + 2  & \text{if } W > 3 \\
				2 & \text{otherwise},
			\end{cases}
		\label{eq:bias_value}
	\end{equation}
	where \(\left\lfloor . \right\rfloor\) denotes the floor function. We aim to recalibrate the attention values of \(T\) proportionally with the expanding context. The optimal \(b\) value was found via greedy search, detailed in Sec.\ref{sec:biases}.

\section{Experiments }\label{sec:experiment}	


\subsection{Dataset \& Models}
We validate MateICL on a range of publicly available datasets covering various NLP tasks, including summarization, commonsense reasoning, sentiment analysis, and reading comprehension, as summarized in Table \ref{tab:datasets}.
As ICL performance is highly dependent on training sample selection \cite{ZhaoWFK021}, following PCW \cite{pcw}, \textbf{we randomly select 30 sets from the training data and report the mean and standard deviation for statistical reliability}. Our experiments were mainly conducted using GPT-2 \cite{radford2019language}, LLaMA \cite{abs-2302-13971}. We have also evaluated on large context models, including Llama-3-8B-Instruct \cite{llama3modelcard}, Qwen2-7B-Instruct \cite{qwen2}, and longchat-7b-v1.5-32k  as well as BLOOM \cite{ScaoWHBBBEMPPRS22}.

\subsection{Prompt Template }
ICL has been shown to be sensitive to the format of prompts \cite{radford2019language}, and to ensure a fair comparison with prior work, we adopt the same prompt template used in their study. Comprehensive details of the adopted template, along with our implementation pseudo-code based on the HuggingFace, are provided in Appendix Section~\ref{sec:setup}.
\begin{table}
	\centering

	\adjustbox{width=\linewidth}{        
		\begin{tabular}{lllll}
			\hline
			\textbf{Task} &\textbf{Dataset} &  \textbf{\# Train} & \textbf{\# Test} & \textbf{Metric} \\
			\hline
			\multirow{1}{*}{Summarize}&
			AGNews  & 120,000 & 7,600 & Accuracy  \\
			\hline
			\multirow{3}{*}{\makecell{Common-\\sense}}&
			COPA  & 400 & 100 & Accuracy  \\
			&HellaSwag  & 39,905 & 10,042 & Accuracy  \\
			&PIQA  & 16,113 & 1,838 & Accuracy  \\
			
			\hline
			
			
			

			\multirow{4}{*}{Sentiment}&
			SST2 & 67,349 & 872 & Accuracy  \\
			&CR&3,394&376&Accuracy\\
			&SUBJ&8,000&2,000&Accuracy\\
			&SST5&8544&1101&Accuracy\\
			\hline
			\multirow{3}{*}{Close QA}&
			ARCE  & 2,241 & 2,365 & Accuracy \\
			&TREC&5452&500&Accuracy\\
			&DBPedia&560,000&70,000&Accuracy\\
			\hline
			\multirow{6}{*}{Reading}&
			OpenBook QA &    4,957 & 500 & Accuracy  \\
			&SQuAD v1  & 87,599 & 10,570 & EM \\
			&WebQS&8,933&2,032&Accuracy\\
			&SQuAD v2&130,319&11,873&EM\\
			&MMLU&99,842&14,042&Accuracy\\
			&StoryCloze&1,871&3,742&Accuracy\\

			\hline
	\end{tabular}}
    
	\label{tab:datasets}
    \caption{Dataset statistics and metrics. }
\end{table}

\subsection{Comparative Baselines}

\begin{table*}[t]
	\centering
	\adjustbox{width=\linewidth}{		
		\begin{tabular}{l|l|c|c|c|ll|ll|ll}  
			\hline					
			\multirow{2}{*}{\makecell[c]{Task}}&	\multirow{2}{*}{\makecell[c]{Dataset}}&\multirow{2}{*}{\# \( \mathcal{L}\)}&	\multirow{2}{*}{\makecell[c]{\(k\)}}&\multirow{2}{*}{\makecell[c]{VanillaICL\\(W=1)}}&\multicolumn{2}{c|}{W=3}&\multicolumn{2}{c|}{W=6} &\multicolumn{2}{c}{W=9} \\\cline{6-11}
			&&&&& \makecell[c]{PCW}& \makecell[c]{MateICL}& \makecell[c]{PCW}& \makecell[c]{MateICL}& \makecell[c]{PCW}& \makecell[c]{MateICL}\\
			\hline

			\hline	

			\multirow{8}{*}{\makecell{GPT2-XL\\ (1.5B)}}
			&SST-2&2&27&90.6$\pm$3.5&92.4$\pm$2.5&\textbf{93.2}$\pm$2.0$^*$&89.4$\pm$3.5&\textbf{92.0}$\pm$2.1&83.7$\pm$1.7&\textbf{92.4}$\pm$2.0\\
			&CR&2&21&79.2$\pm$5.9&81.3$\pm$4.6&\textbf{82.1}$\pm$2.8&\textbf{81.6}$\pm$2.4&81.5$\pm$1.8&82.7$\pm$1.7&\textbf{83.3}$\pm$1.5$^*$\\
			&SUBJ&2&18&68.8$\pm$11.6&64.9$\pm$7.3&\textbf{71.2}$\pm$8.0&57.0$\pm$4.1&\textbf{77.9}$\pm$6.0$^*$&65.6$\pm$3.0&\textbf{75.0}$\pm$5.4\\
			&AGNews&4&11&67.2$\pm$13.2&\textbf{79.6}$\pm$3.4&69.4$\pm$9.7&\textbf{80.4}$\pm$2.3$^*$&65.6$\pm$8.0&\textbf{71.6}$\pm$2.5&67.7$\pm$5.7\\
			&SST5&5&20&38.0$\pm$6.1&41.4$\pm$4.3&\textbf{44.6}$\pm$2.8&38.1$\pm$3.6&\textbf{46.0}$\pm$1.4$^*$&35.3$\pm$2.2&\textbf{46.0}$\pm$1.2\\
			&TREC&6&38&47.9$\pm$5.1&48.7$\pm$2.8&\textbf{49.4}$\pm$3.1&45.5$\pm$2.3&\textbf{50.8}$\pm$3.6&43.1$\pm$1.9&\textbf{51.7}$\pm$2.8$^*$\\
			&DBPedia&14&7&77.5$\pm$9.8&\textbf{87.0}$\pm$4.0&85.9$\pm$4.5&\textbf{88.9}$\pm$3.3$^*$&88.5$\pm$3.4&81.4$\pm$2.1&\textbf{88.8}$\pm$3.0\\
			\cline{2-11}
			&\multicolumn{3}{c|}{Average}&67.0$\pm$7.9&\textbf{70.8}$\pm$4.1&\textbf{70.8}$\pm$4.7&68.7$\pm$3.1&\textbf{71.8}$\pm$3.8&66.2$\pm$2.2&\textbf{72.1}$\pm$3.1\\

			\hline
			\multirow{8}{*}{\makecell{\makecell{LLaMA\\(7B)}}}
			&SST-2&2&48&93.4$\pm$1.3&\textbf{94.9}$\pm$0.6$^*$&94.5$\pm$0.7&91.7$\pm$1.0&\textbf{94.6}$\pm$0.3&84.5$\pm$0.9&\textbf{94.4}$\pm$0.4\\
			&CR&2&39&93.9$\pm$0.7$^*$&93.5$\pm$0.6&\textbf{93.8}$\pm$0.5&90.0$\pm$1.0&\textbf{93.4}$\pm$0.6&79.3$\pm$3.3&\textbf{93.5}$\pm$0.7\\
			&SUBJ&2&32&72.1$\pm$10.0&68.9$\pm$9.5&\textbf{72.5}$\pm$8.2&51.8$\pm$5.1&\textbf{76.9}$\pm$6.2$^*$&48.5$\pm$0.4&\textbf{74.7}$\pm$4.5\\
			&AGNews&4&20&87.9$\pm$2.8&87.3$\pm$1.7&\textbf{88.3}$\pm$1.7&87.4$\pm$1.1&\textbf{88.5}$\pm$1.3$^*$&83.1$\pm$1.8&\textbf{88.5}$\pm$1.1\\
			&SST5&5&36&40.8$\pm$5.6&\textbf{44.6}$\pm$3.8$^*$&42.9$\pm$2.9&40.4$\pm$4.4&\textbf{43.1}$\pm$2.5&22.9$\pm$3.0&\textbf{41.6}$\pm$2.1\\
			&TREC&6&69&83.4$\pm$5.4&81.1$\pm$3.9&\textbf{88.3}$\pm$2.5&55.1$\pm$3.8&\textbf{90.6}$\pm$1.7&41.2$\pm$4.0&\textbf{90.7}$\pm$1.4$^*$\\
			&DBPedia&14&14&86.7$\pm$6.8&\textbf{94.9}$\pm$3.0&94.6$\pm$2.8&95.7$\pm$1.6&\textbf{96.1}$\pm$1.7&92.7$\pm$1.3&\textbf{97.3}$\pm$1.0$^*$\\
			\cline{2-11}
			&\multicolumn{3}{c|}{Average}&79.7$\pm$4.7&80.7$\pm$3.3&\textbf{82.1}$\pm$2.8&73.2$\pm$2.6&\textbf{83.3}$\pm$2.0&64.6$\pm$2.1&\textbf{83.0}$\pm$1.6\\

			\hline
			
			\multirow{8}{*}{\makecell{\makecell{LLaMA\\(30B)}}}
			
			&SST-2&2&48&94.8$\pm$0.5&\textbf{95.0}$\pm$0.7&94.8$\pm$0.5&92.9$\pm$0.5&\textbf{95.0}$\pm$0.2&81.3$\pm$2.2&\textbf{95.3}$\pm$0.3$^*$\\
			&CR&2&39&93.8$\pm$0.5&93.6$\pm$0.5&\textbf{93.7}$\pm$0.6&93.3$\pm$1.1&\textbf{94.4}$\pm$0.4$^*$&87.9$\pm$1.7&\textbf{94.2}$\pm$0.4\\
			&SUBJ&2&32&91.6$\pm$4.2&\textbf{91.6}$\pm$2.2$^*$&89.7$\pm$3.1&85.3$\pm$2.2&\textbf{88.7}$\pm$2.5&57.6$\pm$4.4&\textbf{88.3}$\pm$2.3\\
			&AGNews&4&20&88.0$\pm$4.7&89.4$\pm$0.7&\textbf{90.0}$\pm$0.9&88.0$\pm$0.8&\textbf{90.3}$\pm$0.5$^*$&85.7$\pm$1.9&\textbf{89.6}$\pm$0.7\\
			&SST5&5&36&47.0$\pm$2.6&\textbf{47.5}$\pm$2.3&45.9$\pm$2.1&\textbf{48.4}$\pm$1.0$^*$&46.9$\pm$1.3&40.4$\pm$2.4&\textbf{45.2}$\pm$0.9\\
			&TREC&6&69&87.4$\pm$3.1&90.0$\pm$2.0&\textbf{90.4}$\pm$1.1&67.2$\pm$4.0&\textbf{90.8}$\pm$1.4$^*$&44.0$\pm$8.4&\textbf{90.7}$\pm$0.7\\
			&DBPedia&14&14&90.8$\pm$8.1&\textbf{95.5}$\pm$2.4&95.4$\pm$2.0&\textbf{96.6}$\pm$2.1&96.5$\pm$1.8&96.3$\pm$1.6&\textbf{97.0}$\pm$1.5$^*$\\
			\cline{2-11}
			&\multicolumn{3}{c|}{Average}&84.8$\pm$3.4&\textbf{86.1}$\pm$1.5&85.7$\pm$1.5&81.7$\pm$1.7&\textbf{86.1}$\pm$1.2&70.5$\pm$3.2&\textbf{85.8}$\pm$1.0\\
			
			\hline
		\end{tabular}
	}

	\caption{Averaged scores across text classification datasets. \( \# \mathcal{L} \) denotes the number of classes, and \( k \) represents the number of demonstrations in a single window \( \text{W} \), constrained by the model's capacity (e.g., 1024 tokens for GPT-2). Bold indicates the highest scores, while * denotes statistical significance (t-test, \( p < 0.05 \)). }

	\label{tab:classification}			
	
\end{table*}
\begin{table*}[!t]
	\centering
    	\caption{Comparative results of Multiple Choices Datasets (completion task).  }
	\adjustbox{width=\linewidth}{		
		\begin{tabular}{l|l|c|c|ll|ll|ll|ll}  
			\hline					
			\multirow{2}{*}{Model}&\multirow{2}{*}{Dataset}&\multirow{2}{*}{\makecell[c]{\(k\)}}&\multirow{2}{*}{\makecell[c]{VanillaICL\\W=1}}& \multicolumn{2}{c|}{W=2}&\multicolumn{2}{c|}{W=3} &\multicolumn{2}{c|}{W=4} &\multicolumn{2}{c}{W=6} \\\cline{5-12}
			&&&& \makecell[c]{PCW}& \makecell[c]{MateICL}& \makecell[c]{PCW}& \makecell[c]{MateICL}& \makecell[c]{PCW}& \makecell[c]{MateICL}& \makecell[c]{PCW}& \makecell[c]{MateICL}\\
			\hline	
			
			%
			
			\multirow{8}{*}{GPT-XL}
			&PIQA&12&72.4$\pm$1.1&71.3$\pm$0.9&\textbf{73.4}$\pm$0.8&70.7$\pm$0.8&\textbf{72.6}$\pm$0.7&69.8$\pm$1.0&\textbf{73.8}$\pm$0.7$^*$&68.5$\pm$1.2&\textbf{73.6}$\pm$0.6\\
			&OpenBookAQ&35&34.3$\pm$1.2&34.7$\pm$0.9&\textbf{35.3}$\pm$1.0&33.1$\pm$1.2&\textbf{35.2}$\pm$0.9&32.2$\pm$1.5&\textbf{35.4}$\pm$0.8&29.0$\pm$1.1&\textbf{36.1}$\pm$0.9$^*$\\
			&COPA&47&68.3$\pm$2.1&68.3$\pm$1.8&\textbf{68.7}$\pm$1.4&66.6$\pm$2.2&\textbf{69.1}$\pm$1.7&64.8$\pm$2.3&\textbf{68.2}$\pm$1.3&60.7$\pm$1.5&\textbf{68.6}$\pm$1.3\\
			&HellaSwag&6&46.7$\pm$1.0&44.4$\pm$1.3&\textbf{48.1}$\pm$1.0&43.1$\pm$0.9&\textbf{47.0}$\pm$1.1&41.9$\pm$1.1&\textbf{49.2}$\pm$0.8$^*$&40.9$\pm$0.9&\textbf{48.8}$\pm$0.9\\
			&ARCE&18&50.0$\pm$1.4&49.5$\pm$1.0&\textbf{50.7}$\pm$1.0&48.1$\pm$1.2&\textbf{50.2}$\pm$1.1&47.6$\pm$1.3&\textbf{51.5}$\pm$1.0$^*$&48.3$\pm$1.5&\textbf{50.8}$\pm$1.1\\
			&StoryCloze&13&68.2$\pm$0.9&67.4$\pm$1.3&\textbf{67.7}$\pm$1.1&66.2$\pm$1.0&\textbf{68.1}$\pm$1.0&63.9$\pm$0.9&\textbf{67.9}$\pm$1.0&61.8$\pm$1.2&\textbf{68.5}$\pm$0.8$^*$\\
			&MMLU&2&31.0$\pm$1.3&\textbf{31.3}$\pm$0.9&30.6$\pm$1.2&30.2$\pm$1.5&\textbf{30.5}$\pm$1.1&29.2$\pm$1.5&\textbf{29.3}$\pm$1.0&28.1$\pm$1.3&\textbf{29.0}$\pm$1.1\\
			\cline{2-12}
			&\multicolumn{2}{c|}{Average}&53.0$\pm$1.3&52.4$\pm$1.2&\textbf{53.5}$\pm$1.1&51.1$\pm$1.3&\textbf{53.2}$\pm$1.1&49.9$\pm$1.4&\textbf{53.6}$\pm$0.9&48.2$\pm$1.2&\textbf{53.6}$\pm$1.0\\
			
			\hline

			\multirow{8}{*}{\makecell{LLaMA\\(7B)}}
			&PIQA&23&83.5$\pm$0.7&82.3$\pm$0.6&\textbf{83.9}$\pm$0.5$^*$&81.4$\pm$0.5&\textbf{83.6}$\pm$0.7&80.4$\pm$0.7&\textbf{83.1}$\pm$0.6&79.2$\pm$1.1&\textbf{83.1}$\pm$0.5\\
			&OpenBookAQ&63&50.8$\pm$1.4&\textbf{50.7}$\pm$0.8&50.6$\pm$1.0&48.8$\pm$1.0&\textbf{51.1}$\pm$1.0&47.5$\pm$1.2&\textbf{50.7}$\pm$0.9&40.9$\pm$2.3&\textbf{50.4}$\pm$1.0\\
			&COPA&77&83.0$\pm$2.2&82.5$\pm$1.9&\textbf{83.8}$\pm$1.9&81.4$\pm$1.6&\textbf{83.1}$\pm$2.3&80.1$\pm$1.8&\textbf{83.7}$\pm$0.8&73.6$\pm$2.4&\textbf{85.0}$\pm$1.4\\
			&HellaSwag&12&82.0$\pm${0.9}&\textbf{81.9}$\pm$0.6&81.6$\pm$0.4&81.6$\pm$0.4&\textbf{82.0}$\pm$0.6&81.3$\pm$0.7&\textbf{81.4}$\pm$0.8&80.2$\pm$0.8&\textbf{81.3}$\pm$0.7\\
			&ARCE&33&74.4$\pm$1.3&73.5$\pm$0.8&\textbf{75.2}$\pm$0.8&74.0$\pm$0.9&\textbf{75.1}$\pm$1.1&73.9$\pm$0.9&\textbf{75.8}$\pm$0.7&71.4$\pm$1.5&\textbf{75.8}$\pm$0.7$^*$\\
			&StoryCloze&24&85.0$\pm$0.8&\textbf{85.9}$\pm$0.8&85.8$\pm$1.2&86.2$\pm$0.9&\textbf{86.3}$\pm$0.8&84.7$\pm$0.8&\textbf{85.7}$\pm$0.7&78.5$\pm$1.3&\textbf{86.2}$\pm$0.5\\
			&MMLU&7&41.5$\pm$1.2&\textbf{41.7}$\pm$1.2&41.5$\pm$0.9&\textbf{41.8}$\pm$0.9&41.6$\pm$1.0&\textbf{42.1}$\pm$1.1&40.5$\pm$2.6&\textbf{41.7}$\pm$1.5&38.5$\pm$4.1\\
			\cline{2-12}
			&\multicolumn{2}{c|}{Average}&71.5$\pm$1.2&71.2$\pm$1.0&\textbf{71.8}$\pm$1.0&70.7$\pm$0.9&\textbf{71.8}$\pm$1.1&70.0$\pm$1.0&\textbf{71.6}$\pm$1.0&66.5$\pm$1.6&\textbf{71.5}$\pm$1.3\\
			
			
			\hline
			\multirow{8}{*}{\makecell{LLaMA\\(30B)}}
			&PIQA&23&84.1$\pm$0.6&83.6$\pm$0.6&\textbf{83.6}$\pm$0.6&83.9$\pm$0.4&\textbf{84.0}$\pm$0.6&83.3$\pm$0.5&\textbf{84.0}$\pm$0.6&82.8$\pm$0.8&\textbf{84.0}$\pm$0.5\\
			&OpenBookAQ&63&53.9$\pm$1.3&54.0$\pm$1.4&\textbf{54.4}$\pm$1.1&53.5$\pm$0.5&\textbf{55.0}$\pm$1.1&52.9$\pm$1.0&\textbf{54.8}$\pm$1.0&48.8$\pm$1.8&\textbf{54.7}$\pm$1.1\\
			&COPA&77&87.6$\pm$2.1&85.9$\pm$2.2&\textbf{88.2}$\pm$2.0&86.1$\pm$2.0&\textbf{86.9}$\pm$2.3&85.2$\pm$2.2&\textbf{87.8}$\pm$2.2&76.6$\pm$2.8&\textbf{88.1}$\pm$1.5\\
			&HellaSwag&12&88.4$\pm$0.8&88.4$\pm$0.4&\textbf{88.6}$\pm$0.4&88.5$\pm$0.7&\textbf{88.6}$\pm$0.6&87.9$\pm$0.6&\textbf{88.2}$\pm$0.8&\textbf{87.6}$\pm$0.8&87.4$\pm$0.9\\
			&ARCE&33&83.9$\pm$0.8&\textbf{85.2}$\pm$0.8$^*$&84.8$\pm$0.7&84.8$\pm$0.8&\textbf{85.0}$\pm$0.7$^*$&84.2$\pm$0.7&\textbf{84.9}$\pm$1.2$^*$&79.4$\pm$0.7&\textbf{84.9}$\pm$0.7$^*$\\
			&StoryCloze&24&86.4$\pm$0.7&\textbf{86.2}$\pm$1.0&86.0$\pm$0.6&86.5$\pm$0.6&\textbf{86.8}$\pm$0.6&86.1$\pm$0.5&\textbf{86.3}$\pm$0.6&83.5$\pm$1.2&\textbf{86.0}$\pm$0.5\\
			&MMLU&7&48.0$\pm$1.2&48.7$\pm$1.1&\textbf{49.6}$\pm$1.3$^*$&48.2$\pm$1.1&\textbf{49.4}$\pm$1.2$^*$&48.2$\pm$1.0&\textbf{49.0}$\pm$1.1$^*$&46.9$\pm$0.9&\textbf{49.0}$\pm$1.0$^*$\\
			\cline{2-12}
			&\multicolumn{2}{c|}{Average}&76.0$\pm$1.1&76.0$\pm$1.1&\textbf{76.5}$\pm$1.0&75.9$\pm$0.9&\textbf{76.5}$\pm$1.0&75.4$\pm$0.9&\textbf{76.4}$\pm$1.1&72.2$\pm$1.3&\textbf{76.3}$\pm$0.9\\
			
			\hline	
			
		\end{tabular}
	}	 		

	\label{tab:multi_choice}	
    \caption{Comparative results of Multiple Choices Datasets (completion task).  }
\end{table*}

\begin{table*}[ht]
	\centering
	\adjustbox{width=\linewidth}{		
		\begin{tabular}{l|cccccc|cc|c}  
			\hline	
			\multirow{2}{*}{Task}& \multicolumn{6}{c|}{Learning-free }	& \multicolumn{2}{c|}{Learning-based }&\multirow{2}{*}{MateICL}\\\cline{2-9}
			&Zero & Random & Kmeans & BM25 & E5base & SBERT & EPR & LLM-R&\\
			\hline			
			AGNews&31.5&67.4&71.9&90.6&90.6&90.2&91.8&\textbf{93.1}$\pm$0.5&87.4$\pm$0.9\\
			ARC Chall.&35.6&39.7&40.5&40.3&44.6&42.8&43.0&43.7$\pm$0.2&\textbf{47.3}$\pm$0.7\\
			ARC Easy&51.0&60.0&61.8&59.9&63.0&63.1&63.1&63.5$\pm$0.1&\textbf{75.8}$\pm$0.1\\
			BoolQ&64.7&70.0&69.0&74.7&72.4&73.9&74.8&74.9$\pm$0.6&\textbf{76.8}$\pm$1.0\\
			COPA&66.0&80.0&85.0&78.0&83.0&82.0&82.0&\textbf{84.0}$\pm$0.0&{83.8}$\pm$0.8\\
			E2E NLG&34.6&52.7&46.4&54.5&51.8&50.2&53.6&\textbf{54.8}$\pm$0.1&54.4$\pm$0.8\\
			Gigaword&15.3&30.0&30.7&32.7&32.5&32.6&32.4&32.5$\pm$0.6&\textbf{38.6}$\pm$0.2\\
			HellaSwag&71.5&73.9&74.0&74.9&75.2&75.3&75.2&\textbf{75.4}$\pm$0.0&75.1$\pm$0.5\\
			MRPC&69.1&49.5&38.0&61.8&41.2&52.7&55.9&\textbf{71.9}$\pm$6.9&62.3$\pm$8.2\\
			MultiRC&57.0&48.5&34.1&54.2&56.0&55.3&50.4&52.2$\pm$0.6&\textbf{59.8}$\pm$3.2\\
			NQ&0.3&21.5&22.6&37.6&39.3&39.4&39.2&\textbf{39.2}$\pm$0.1&37.2$\pm$0.1\\
			OpenBook QA&41.6&49.8&49.0&49.6&51.4&51.4&49.6&52.1$\pm$1.1&\textbf{61.8}$\pm$0.7\\
			PIQA&77.0&79.1&79.4&81.3&81.3&80.7&80.5&\textbf{81.0}$\pm$0.4&80.2$\pm$0.1\\
			RTE&59.6&59.9&58.5&65.7&63.9&67.2&66.8&68.7$\pm$1.3&\textbf{76.8}$\pm$1.6\\
			Sentiment140&49.3&88.6&89.4&90.8&93.9&92.2&91.4&90.7$\pm$0.3&\textbf{93.6}$\pm$0.1\\
			SQuAD v1&2.1&64.1&62.3&61.2&60.8&61.6&64.3&56.8$\pm$3.4&\textbf{70.0}$\pm$0.3\\
			SST2&54.4&85.9&89.7&84.4&92.1&87.6&88.7&\textbf{93.6}$\pm$0.4&92.4$\pm$2.4\\
			Winogrande&62.0&66.7&66.5&67.5&66.9&66.5&66.5&\textbf{67.7}$\pm$0.4&66.6$\pm$0.6\\
			WSC&64.4&60.6&56.7&56.7&61.5&63.5&61.5&\textbf{63.5}$\pm$2.4&61.6$\pm$4.7\\
			Yelp&47.9&92.0&93.5&93.5&97.3&95.9&95.1&95.7$\pm$0.2&\textbf{97.8}$\pm$0.1\\
			\hline			
		\end{tabular}
	}		
    \caption{Comparative results with retrieval-based baselines on LLaMA-7B.  MateICL achieves this performance while  eliminating the complexity introduced by external encoders, e.g., SBERT.}
	
    \label{tab:compare_to_retreivale}	

\end{table*}

We conduct comparative experiments to these approaches: 

\begin{itemize}
	\item \textbf{Context Extension-Based Approaches:}
	These methods do not require further training:

	 \textbf{VanillaICL}: A standard ICL approach that concatenates all demonstrations of a single context window.
	 \textbf{PCW} \cite{pcw}: Adjusts position encoding and attention masks to facilitate multiple context windows without additional training.
	 Recent strategies \cite{qwen2} have enabled LLMs to process up to 32k tokens. To evaluate performance under limited GPU resources, we introduce a baseline, \textbf{InfICL}, which is the same idea of MateICL without attention calibration, rather than concatenating all demonstrations together.
	 \textbf{StructuredICL} \cite{Yaru-2212-06713}: Enhances non-learned positional encoding LLMs through rescaled attention.

	\item \textbf{Retrieval Free-Learning Methods}: \textbf{Random}: Random retrieval of context examples. \textbf{K-means}: Clustered Search Accelerator for grouping similar context examples.
			 \textbf{BM25}: A classic retrieval model based on term frequency and inverse document frequency;
			 \textbf{E5$_{\text{base}}$} \cite{wang2022simlm}: A model that performs retrieval based on embedding similarity.
			 \textbf{SBERT} \cite{reimers2019sentence}: Uses sentence embeddings for retrieval.

		\item \textbf{Retrieval Learning-Based Methods}:
		 \textbf{EPR} \cite{RubinHB22}: A method that retrieves the most relevant singleton in-context example by selecting the Top-K most similar examples during inference.
		 \textbf{LLM-R} \cite{llm-r}: Proposes an iterative approach that trains a reward model with LLM feedback to assess the quality of candidate examples, followed by knowledge distillation to train a bi-encoder-based dense retriever.

\end{itemize}
\begin{table*}[!t]
	\centering
		\adjustbox{width=\linewidth}{		
		\begin{tabular}{c|l|cc|cc|cc}  
			\hline	
			\multirow{2}{*}{Task}&	\multirow{2}{*}{\makecell[c]{Dataset}}&\multicolumn{2}{c|}{Qwen2-7B-Instruct}&\multicolumn{2}{c|}{Llama-3-8B-Instruct}&\multicolumn{2}{c}{Longchat-7B-v1.5}\\\cline{3-8}
			&&InfICL& MateICL&InfICL& MateIC&InfICL& MateIC\\
			\hline
			\multirow{8}{*}{\makecell{Text\\Classification}}
&SST-2		&\textbf{96.7}$\pm$0.5&95.8$\pm$0.5&94.2$\pm$0.5&\textbf{95.0}$\pm$0.5&85.6$\pm$7.5&\textbf{95.8}$\pm$0.7\\
&CR		&\textbf{92.6}$\pm$1.3&92.2$\pm$0.7&\textbf{92.4}$\pm$0.6&91.8$\pm$1.4&84.8$\pm$7.0&\textbf{93.6}$\pm$0.8\\
&SUBJ		&\textbf{94.6}$\pm$1.1&94.3$\pm$0.5&93.6$\pm$1.0&\textbf{95.8}$\pm$0.8&81.7$\pm$3.4&\textbf{88.6}$\pm$2.0\\
&AGNews		&82.4$\pm$1.1&\textbf{83.8}$\pm$1.4&87.8$\pm$0.9&\textbf{89.0}$\pm$1.4&87.7$\pm$0.7&\textbf{88.2}$\pm$0.7\\
&SST5		&47.8$\pm$1.6&\textbf{50.3}$\pm$2.0&40.6$\pm$2.0&\textbf{43.6}$\pm$1.2&43.0$\pm$5.2&\textbf{54.0}$\pm$1.5\\
&TREC		&\textbf{90.2}$\pm$1.0&90.1$\pm$1.2&87.4$\pm$2.4&\textbf{90.0}$\pm$2.5&72.2$\pm$1.7&\textbf{89.4}$\pm$2.2\\
&DBPedia	&\textbf{97.8}$\pm$0.2&97.6$\pm$0.7&{98.6}$\pm$0.5&\textbf{98.8}$\pm$0.4&96.4$\pm$1.3&\textbf{97.8}$\pm$0.3\\\cline{2-8}

&Average&86.0$\pm$1.0&\textbf{86.3}$\pm$1.0&84.9$\pm$1.1&\textbf{86.3}$\pm$1.2&78.8$\pm$3.8&\textbf{86.8}$\pm$1.2\\

\hline
\multirow{7}{*}{\makecell{Completion\\Task}}
&PIQA		&\textbf{79.8}$\pm$0.3&79.5$\pm$0.9&74.1$\pm$0.3&\textbf{75.4}$\pm$0.4&73.4$\pm$1.7&\textbf{81.9}$\pm$0.5\\
&OpenBookAQ&48.4$\pm$0.6&\textbf{48.6}$\pm$0.7&48.2$\pm$1.0&\textbf{49.0}$\pm$0.6&40.3$\pm$1.6&\textbf{50.7}$\pm$0.9\\
&COPA		&78.9$\pm$1.1&\textbf{81.0}$\pm$0.8&76.8$\pm$0.5&\textbf{79.7}$\pm$0.4&63.8$\pm$4.8&\textbf{84.2}$\pm$0.2\\
&HellaSwag	&74.2$\pm$0.5&\textbf{75.6}$\pm$0.8&74.9$\pm$0.3&\textbf{77.2}$\pm$0.7&66.7$\pm$2.3&\textbf{80.0}$\pm$0.4\\
&ARCE		&\textbf{70.3}$\pm$0.9&70.2$\pm$0.9&68.4$\pm$0.8&\textbf{69.8}$\pm$0.4&70.9$\pm$1.2&\textbf{77.3}$\pm$0.5\\
&StoryCloze&82.2$\pm$0.9&\textbf{83.0}$\pm$0.5&81.6$\pm$1.3&\textbf{83.3}$\pm$1.2&75.4$\pm$0.7&\textbf{83.8}$\pm$1.0\\

\cline{2-8}
&Average&72.3$\pm$0.7&\textbf{73.0}$\pm$0.8&70.7$\pm$0.7&\textbf{72.4}$\pm$0.6&65.1$\pm$2.0&\textbf{76.3}$\pm$0.6\\

			\hline			
		\end{tabular}
		}	
          \caption{Comparative evaluation on large-scaled context models. }

	\label{tab:compare_to_32k}	
\end{table*}

\begin{table*}
	\centering
	\adjustbox{width=\linewidth}{		
		\begin{tabular}{l|l|c|c|ll|ll|ll}  
			\hline					
			\multirow{2}{*}{Model}&\multirow{2}{*}{Dataset}&\multirow{2}{*}{\makecell[c]{\(k\) instances\\per W}}&\multirow{2}{*}{\makecell[c]{ICL\\W=1}}&\multicolumn{2}{c|}{W=2}&\multicolumn{2}{c|}{W=3} &\multicolumn{2}{c}{W=4} \\\cline{5-10}
			&&&& \makecell[c]{PCW}& \makecell[c]{MateICL}& \makecell[c]{PCW}& \makecell[c]{MateICL}& \makecell[c]{PCW}& \makecell[c]{MateICL}\\

			\hline		
			\multirow{4}{*}{GPT-Large}&SQuAD&2&41.2$\pm$2.0$^*$&39.9$\pm$2.1&\textbf{41.0}$\pm$2.0&35.9$\pm$1.6&\textbf{39.2}$\pm$1.9&32.6$\pm$2.1&\textbf{37.8}$\pm$2.2\\
			&SQuADV2&2&48.1$\pm$0.2&48.2$\pm$0.6&\textbf{48.2}$\pm$0.8&\textbf{48.2}$\pm$0.4&48.1$\pm$0.1&\textbf{48.1}$\pm$0.2&48.0$\pm$0.1\\
			&WebQS&43&8.9$\pm$1.1$^*$&8.2$\pm$1.0&\textbf{8.5}$\pm$1.0&7.7$\pm$1.4&\textbf{8.1}$\pm$1.4&7.0$\pm$1.1&\textbf{7.7}$\pm$0.9\\
			\cline{2-10}
			&\multicolumn{2}{c|}{Average}&32.7$\pm$1.1&32.1$\pm$1.2&\textbf{32.6}$\pm$1.3&30.6$\pm$1.1&\textbf{31.8}$\pm$1.1&29.2$\pm$1.1&\textbf{31.2}$\pm$1.1\\
			
			\hline	
			
			\multirow{4}{*}{GPT-XL}&SQuAD&2&44.0$\pm$4.5&44.1$\pm$3.0&\textbf{45.2}$\pm$3.1$^*$&40.3$\pm$2.5&\textbf{42.6}$\pm$3.0&38.0$\pm$2.6&\textbf{41.7}$\pm$2.5\\
			&SQuADV2&2&48.2$\pm$0.3&48.6$\pm$0.9&\textbf{48.7}$\pm$1.1&48.6$\pm$1.0&\textbf{48.7}$\pm$1.1&48.7$\pm$0.9&\textbf{48.7}$\pm$0.8\\
			&WebQS&43&13.2$\pm$0.9&13.1$\pm$0.8&\textbf{13.2}$\pm$0.7$^*$&12.2$\pm$0.9&\textbf{12.9}$\pm$0.9&11.1$\pm$1.0&\textbf{12.3}$\pm$0.9\\
			\cline{2-10}
			&\multicolumn{2}{c|}{Average}&35.1$\pm$1.9&35.3$\pm$1.6&\textbf{35.7}$\pm$1.6&33.7$\pm$1.5&\textbf{34.7}$\pm$1.7&32.6$\pm$1.5&\textbf{34.2}$\pm$1.4\\
			
			\hline	
			
			\multirow{4}{*}{LLaMA-7B}&SQuAD&4&79.7$\pm$2.1&80.0$\pm$1.6&\textbf{80.5}$\pm$1.7&79.5$\pm$1.7&\textbf{80.6}$\pm$1.3$^*$&78.5$\pm$1.7&\textbf{80.4}$\pm$1.4\\
			&SQuADV2&4&57.9$\pm$6.5&57.4$\pm$6.7&\textbf{57.6}$\pm$7.0&55.5$\pm$4.4&\textbf{56.2}$\pm$4.9&55.5$\pm$4.2&\textbf{55.9}$\pm$4.5\\
			&WebQS&72&36.4$\pm$2.4&\textbf{37.1}$\pm$1.6$^*$&36.9$\pm$1.5&34.8$\pm$1.6&\textbf{36.8}$\pm$1.4&33.9$\pm$1.4&\textbf{35.0}$\pm$1.3\\
			\cline{2-10}
			&\multicolumn{2}{c|}{Average}&58.0$\pm$3.7&58.2$\pm$3.3&\textbf{58.3}$\pm$3.4&56.6$\pm$2.6&\textbf{57.9}$\pm$2.5&56.0$\pm$2.4&\textbf{57.1}$\pm$2.4\\
			
			\hline
			
			\multirow{4}{*}{LLaMA-13B}&SQuAD&4&75.7$\pm$2.9&\textbf{78.0}$\pm$2.3&77.8$\pm$2.3&78.2$\pm$2.0&\textbf{78.8}$\pm$1.8$^*$&77.4$\pm$1.8&\textbf{78.5}$\pm$1.9\\
			&SQuADV2&4&58.7$\pm$7.5&65.5$\pm$7.7&\textbf{65.6}$\pm$7.9$^*$&60.8$\pm$7.8&\textbf{60.9}$\pm$7.8&65.5$\pm$7.7&\textbf{65.6}$\pm$7.6\\
			&WebQS&72&39.6$\pm$2.3&40.7$\pm$1.8&\textbf{41.2}$\pm$1.7$^*$&39.2$\pm$1.4&\textbf{40.7}$\pm$1.4&37.8$\pm$1.4&\textbf{40.0}$\pm$1.7\\
			\cline{2-10}
			&\multicolumn{2}{c|}{Average}&58.0$\pm$4.2&61.4$\pm$3.9&\textbf{61.5}$\pm$4.0&59.4$\pm$3.7&\textbf{60.1}$\pm$3.7&60.2$\pm$3.6&\textbf{61.4}$\pm$3.7\\
			
			\hline	
			
			\multirow{4}{*}{LLaMA-30B}&SQuAD&4&83.7$\pm$2.4&84.3$\pm$1.2&\textbf{84.5}$\pm$1.3&83.8$\pm$1.4&\textbf{84.6}$\pm$1.3&84.8$\pm$1.1&\textbf{85.9}$\pm$1.2$^*$\\
			&SQuADV2&4&57.4$\pm$6.1&\textbf{58.6}$\pm$5.6&58.5$\pm$5.6&56.8$\pm$4.3&\textbf{56.9}$\pm$4.3&59.6$\pm$5.3&\textbf{59.8}$\pm$5.5\\
			&WebQS&72&42.9$\pm$2.4$^*$&41.8$\pm$1.4&\textbf{42.8}$\pm$1.3&\textbf{42.2}$\pm$1.4&41.6$\pm$1.0&38.2$\pm$1.4&\textbf{41.8}$\pm$1.3\\
			
			\cline{2-10}
			&\multicolumn{2}{c|}{Average}&61.3$\pm$3.6&61.6$\pm$2.7&\textbf{61.9}$\pm$2.7&60.9$\pm$2.4&\textbf{61.0}$\pm$2.2&60.9$\pm$2.6&\textbf{62.5}$\pm$2.7\\
			
			\hline

	\end{tabular}
    }			
        \caption{The comparative scores of Machine Reading Comprehension Datasets (Generating Task). }
	\label{tab:generation}	
\end{table*}

\begin{table*}[t]
	\centering
	\adjustbox{width=\linewidth}{		
		\begin{tabular}{c|l|c|c|c|ll|ll|ll}  
			\hline					
			\multirow{2}{*}{\makecell[c]{Task}}&\multirow{2}{*}{\makecell[c]{Dataset}}&\multirow{2}{*}{\# Labels}&\multirow{2}{*}{\makecell[c]{\(k\) instances\\per W}}&\multirow{2}{*}{\makecell[c]{ICL\\(W=1)}}&\multicolumn{2}{c|}{W=3}&\multicolumn{2}{c|}{W=9} &\multicolumn{2}{c}{W=18} \\\cline{6-11}
			&&&&& \makecell[c]{StructedICL}& \makecell[c]{MateICL}& \makecell[c]{StructedICL}& \makecell[c]{MateICL}& \makecell[c]{StructedICL}& \makecell[c]{MateICL}\\
			\hline
			
			\multirow{7}{*}{\makecell{Text\\Classification}}&SST-2 &2&48& 78.2 $\pm$ 7.2&84.2$\pm$2.9&\textbf{87.1}$\pm$1.9&86.6$\pm$1.5&\textbf{88.8}$\pm$0.9$^*$&85.6$\pm$0.7&\textbf{87.8}$\pm$0.4 \\
			&SUBJ &2&32& 85.0 $\pm$ 3.8&89.4$\pm$2.7&\textbf{91.7}$\pm$2.9&90.8$\pm$0.4&\textbf{93.1}$\pm$0.6$^*$&89.9$\pm$1.1&\textbf{92.8}$\pm$0.9 \\
			&AGNews&4&20& 81.0 $\pm$ 3.6&84.2$\pm$2.9&\textbf{87.0}$\pm$2.9&84.7$\pm$2.2&\textbf{87.6}$\pm$2.0&84.5$\pm$1.0&\textbf{87.8}$\pm$0.8$^*$ \\
			&SST5 &5&36& 49.0 $\pm$ 2.3&51.2$\pm$1.0&\textbf{53.9}$\pm$1.3&53.4$\pm$1.2&\textbf{56.2}$\pm$2.0$^*$&52.4$\pm$1.4&\textbf{54.8}$\pm$1.1 \\
			&TREC &6&69& 47.4 $\pm$ 6.3&49.0$\pm$3.6&\textbf{49.5}$\pm$3.2&52.5$\pm$2.4&\textbf{54.7}$\pm$1.2&54.4$\pm$1.2&\textbf{57.7}$\pm$1.5$^*$ \\
			&DBPedia &14&14& 91.0 $\pm$ 1.4&92.8$\pm$1.3&\textbf{95.0}$\pm$1.7$^*$&92.4$\pm$1.3&\textbf{95.0}$\pm$1.4&92.0$\pm$0.6&\textbf{94.2}$\pm$0.6 \\
			\cline{2-11}
			&\multicolumn{3}{c|}{Average}&71.9$\pm$4.1&75.1$\pm$2.4&\textbf{77.4}$\pm$2.3&76.7$\pm$1.5&\textbf{79.2}$\pm$1.3&76.5$\pm$1.0&\textbf{79.2}$\pm$0.9\\
			
			\hline	
			\multirow{8}{*}{NLI}&Wic &2&39& 50.0 $\pm$ 2.1&49.8$\pm$0.9&\textbf{52.8}$\pm$2.7$^*$&51.4$\pm$2.0&\textbf{52.0}$\pm$1.6&50.3$\pm$0.9&\textbf{52.4}$\pm$0.7 \\
			&BoolQ &2&5& 51.1 $\pm$ 7.5&50.5$\pm$2.1&\textbf{56.0}$\pm$6.9&49.1$\pm$1.9&\textbf{54.5}$\pm$2.6&53.8$\pm$4.4&\textbf{56.5}$\pm$3.2$^*$ \\
			&MultiRC &2&2& 56.8 $\pm$ 2.2&57.0$\pm$2.2&\textbf{58.8}$\pm$3.7&57.1$\pm$0.6&\textbf{60.2}$\pm$0.6$^*$&57.3$\pm$0.6&\textbf{59.6}$\pm$1.1 \\
			&RTE &2&10& 53.4 $\pm$ 3.0&51.3$\pm$2.3&\textbf{52.4}$\pm$1.1&52.3$\pm$2.1&\textbf{53.6}$\pm$1.3&50.2$\pm$2.0&\textbf{52.9}$\pm$1.4 \\
			&QNLI &2&20& 54.9 $\pm$ 3.6&55.1$\pm$1.0&\textbf{57.5}$\pm$2.8&55.7$\pm$1.3&\textbf{58.9}$\pm$1.8$^*$&54.0$\pm$1.7&\textbf{57.2}$\pm$0.9 \\
			&SICK &3&36& 55.3 $\pm$ 6.6&57.6$\pm$2.0&\textbf{61.7}$\pm$1.5$^*$&58.5$\pm$1.6&\textbf{60.0}$\pm$0.7&57.2$\pm$1.0&\textbf{59.6}$\pm$0.9 \\
			&SNLI &3&35& 36.6 $\pm$ 4.6&41.1$\pm$3.6&\textbf{42.8}$\pm$3.0&41.7$\pm$3.0&\textbf{43.8}$\pm$3.4&38.4$\pm$1.4&\textbf{40.9}$\pm$1.9 \\
			\cline{2-11}
			&\multicolumn{3}{c|}{Average}&51.2$\pm$4.2&51.8$\pm$2.0&\textbf{54.6}$\pm$3.1&52.3$\pm$1.8&\textbf{54.7}$\pm$1.7&51.6$\pm$1.7&\textbf{54.2}$\pm$1.4\\
			
			\hline	
		\end{tabular}
	}	
	 \caption{Our MateICL against Structure Prompting (StructedICL) on the Classification Datasets  using BLOOM-7B1. Note that we expend the context window (\(W=18\)). Bold indicates highest scores; * denotes statistical significance (t-test, \(p < 0.05\)). }
	\label{tab:nli_bloom-7b1}	
\end{table*}

\section{ Empirical Evaluation }\label{sec:results}	
We evaluate MateICL based on the criterion: \textit{Does scaling the context enhance ICL performance?}


\subsection{Can Scaling Context Enhance ICL?}

The ultimate goal of our proposal is to enable existing LLMs to scale their context in order to enhance ICL performance. To evaluate this, we conducted extensive experiments on text classification and completion tasks. For each task, we included up to 9 context windows (\(\text{W}\)), each containing \(k\) instances, within the context limits of the LLM (e.g., GPT-2 allows up to 27 demonstrations per window for SST2). We report the results in Tables \ref{tab:classification},  \ref{tab:multi_choice}, and \ref{tab:generation} for text classification, multiple choice and generative tasks, respectively. From our results, we have made the following observations:  
(1) \textbf{VanillaICL}, with a limited number of demonstrations, reports modest performance compared to both PCW and MateICL across all tasks in terms of accuracy and stability. This aligns with our claim that extending the context is necessary for improving performance.  
(2) Extending the context in \textbf{PCW} indeed leads to significant performance improvements; however, this gain is limited to specific context window sizes, particularly with \(\text{W} = 3\), after which performance begins to decline as \(\text{W}\) increases. This trend is expected, as discussed in Eq. \ref{eq:newproof}, where \(\nu(\mathbf{x}_r)\) increases with the number of demonstrations, causing the model to focus less on the query components. This effect is further exacerbated by the absence of mechanisms that encourage the model to attend to the query tokens.  
(3)\textbf{ MateICL shows significant improvements over both \textbf{VanillaICL} and \textbf{PCW} across all tasks. Moreover, when scaling the context, MateICL demonstrates that, in cases where no improvement in accuracy is observed, performance stability is maintained, and even improved, without a drop in performance.  }
(4) While {MateICL} generally outperforms PCW in multiple-choice tasks, the improvement margin is small, especially when compared to classification tasks, which aligns with the findings in \cite{radford2019language}.

\subsection{Context Extension vs. Context Retrieval}
Instead of concatenating demonstrations to form the context, retrieval-based approaches first select the most relevant examples for the in-task query \cite{icl_survey_2024}, and then use them as context for the final inference. \textit{However, these approaches introduce two types of complexity, which our MateICL addresses.} The need to fine-tune an external scorer and the requirement to infer the query twice (i.e., selecting the top relevant examples and performing final inference) \cite{nbce}. We conduct experiments against both dense and sparse retrieval models, using the same settings as LLM-R \cite{llm-r} with 8 shots and the same ICL pool for consistency. The results, shown in Table \ref{tab:compare_to_retreivale}, reveal that large context does not always benefit tasks like text classification (e.g., Agnews and Sentiment). Compared to the learning-free baseline, MateICL significantly outperforms  without the need for an external scorer. When compared to learning-based baselines, MateICL performs better in some tasks and is highly competitive in others. It is important to note that fine-tuning a scorer is time-consuming and not practical in many scenarios.
\textbf{Notably, MateICL achieves this performance while significantly reducing complexity from \(O(n \times k)\), where \(k\) represents the top shots, to \(O(n/m)\), with \(m\) being the number of windows. This reduction also accounts for the complexity introduced by external encoders, e.g., SBERT.}
\subsection{MateICL is Effective in Limited GPUs}

While recent models allow concatenating a large number of demonstrations, they require substantial GPU memory \cite{qwen2}. When dealing with numerous examples, humans typically draw insights from separate examples and integrate them, avoiding distractions from focusing on too many examples at once. Motivated by this, instead of concatenating all demonstrations and performing inference in one step, we propose dividing the context into multiple windows (i.e., 4096 tokens per window), each processed separately. This approach allows large context inclusion even with limited GPU memory. We name this strategy InfICL and compare it with our MateICL. Both approaches use the same settings, including the number of demonstrations per window and the total number of windows.
\textbf{Results in Table \ref{tab:compare_to_32k} show that  MateICL enables models with lower performance to surpass even the most powerful models. For instance,  Longchat-7B-v1.5 under MateICL outperforms instruction-based models.} 
Compared to InfICL, MateICL achieves significant improvements of up to 8\% and 11.2\% in text and completion tasks, respectively.
 MateICL outperforms InfICL on average across all tasks and models. For Longchat-7B-v1.5, MateICL achieves significant improvements of up to 8\% and 11.2\% in text and completion tasks, respectively. 
However, the performance gains are more modest for the instruction-tuned models, Qwen2-7B-Instruct and LLaMA-3-8B-Instruct. This outcome is expected, as these models have already undergone supervised fine-tuning on diverse instruction-following datasets, which likely equips them with stronger generalization capabilities and reduces their dependence on explicit demonstration selection during inference.

\begin{table*}[!ht]
	\centering
	\adjustbox{width=\linewidth}{		
		\begin{tabular}{l|c|c|ll|ll|ll|ll}  
			\hline					
			\multirow{2}{*}{Dataset}&\multirow{2}{*}{\makecell[c]{\(k\) instances\\per W}}&\multirow{2}{*}{\makecell[c]{ICL\\W=1}}&\multicolumn{2}{c|}{W=2}&\multicolumn{2}{c|}{W=3} &\multicolumn{2}{c|}{W=4} &\multicolumn{2}{c}{W=6} \\\cline{4-11}
			&&& \makecell[c]{StructedICL}& \makecell[c]{MateICL}& \makecell[c]{StructedICL}& \makecell[c]{MateICL}& \makecell[c]{StructedICL}& \makecell[c]{MateICL}& \makecell[c]{StructedICL}& \makecell[c]{MateICL}\\
			\hline			
			PIQA &23& 71.2 $\pm$ 0.9&71.3$\pm$1.5&\textbf{72.3}$\pm$1.2&72.2$\pm$1.7&\textbf{73.1}$\pm$1.7&72.2$\pm$0.8&\textbf{72.8}$\pm$1.1&72.2$\pm$1.2&\textbf{72.7}$\pm$1.0 \\
			OpenBookAQ &63& 31.6 $\pm$ 1.2&31.8$\pm$1.0&\textbf{33.3}$\pm$0.7$^*$&32.0$\pm$1.3&\textbf{32.6}$\pm$1.1&32.0$\pm$0.8&\textbf{32.5}$\pm$1.9&32.5$\pm$1.4&\textbf{33.0}$\pm$1.7 \\
			COPA &77& 72.2 $\pm$ 2.0&73.8$\pm$2.8&\textbf{74.8}$\pm$2.8&73.7$\pm$2.2&\textbf{75.3}$\pm$1.7&72.9$\pm$2.8&\textbf{75.1}$\pm$2.6&74.1$\pm$3.1&\textbf{74.4}$\pm$2.0 \\
			HellaSwag &12& 57.0 $\pm$ 0.9&57.4$\pm$0.8&\textbf{58.2}$\pm$1.0$^*$&57.0$\pm$0.6&\textbf{57.4}$\pm$1.0&56.8$\pm$1.0&\textbf{57.0}$\pm$1.0&56.8$\pm$1.2&\textbf{57.7}$\pm$1.5 \\
			ARCE &33& 64.2 $\pm$ 1.5&65.0$\pm$1.4&\textbf{65.6}$\pm$1.1&64.8$\pm$1.5&\textbf{65.6}$\pm$1.1&65.0$\pm$0.9&\textbf{66.3}$\pm$0.9&65.8$\pm$1.1&\textbf{66.1}$\pm$0.8 \\
			
			\hline
			\multicolumn{2}{c|}{Average}&59.2$\pm$1.3&59.9$\pm$1.5&\textbf{60.8}$\pm$1.4&59.9$\pm$1.5&\textbf{60.8}$\pm$1.3&59.8$\pm$1.3&\textbf{60.7}$\pm$1.5&60.3$\pm$1.6&\textbf{60.8}$\pm$1.4\\

			\hline
		\end{tabular}
	}	
      \caption{Our MateICL against Structure Prompting (StructedICL) on the Multiple Choices Datasets (completion task)  using BLOOM-7B1. Best scores are highlighted in bold. }
	\label{tab:multi_choice_bloom}	
\end{table*}
\begin{figure*}
	\centering
	\includegraphics[scale=0.8]{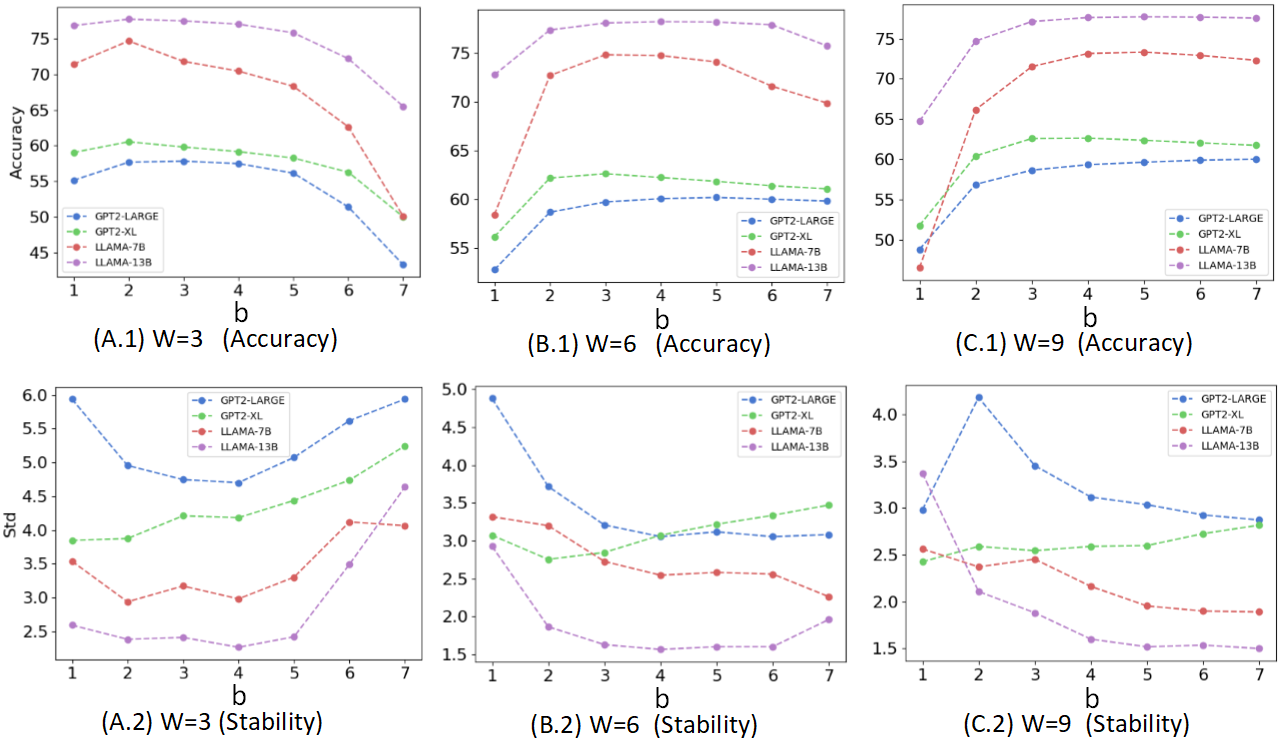}
	\caption{Averaged accuracy and standard deviation across text classification datasets of  (\(1 \leq b \leq 7\)) of Eq.\ref{eq:bias_value} within context windows (\(W=6\)) and (\(W=9\)).}
	\label{fig:various_bias}
\end{figure*}

\subsection{Machine Reading Comprehension Task}\label{sec:mrc_task}
The machine reading comprehension task, often termed as free-form completion, is designed to evaluate LLMs' capabilities in text comprehension. This is achieved by providing the LLMs with a text passage and a related question, thereby evaluating their abilities of comprehension of the text by extracting necessary information and generating an accurate and context-relevant answers.
We follow \cite{RaffelSRLNMZLL20,BrownMRSKDNSSAA20}, employing beam search with a beam width of  4 and a length penalty of \(\alpha=0.6\). We use the F1 for SQuAD and SQuADV2 and accuracy for WebQS. 

The results in Table \ref{tab:generation}  suggests improvements depend on model size, with larger models showing better performance. For LLaMA-30B, improvements over LLaMA-7B in ICL, PCW by \(3.9\%\). Consistent with findings in \cite{BrownMRSKDNSSAA20}, more demonstrations do not always enhance performance, possibly due to diluted supervision and the lack of generation constraints in this task.

\subsection{MateICL against StructedICL}\label{sec:struct}

To evaluate our MateICL against Structure Prompting (StructedICL) \cite{Yaru-2212-06713}, we conducted additional experiments on BLOOM-7B1, a non-learned positional encoding (e.g., ALiBi-based \cite{PressSL22}). Performance results are reported in Tables \ref{tab:nli_bloom-7b1} for text classification and NLI tasks, and Table \ref{tab:multi_choice_bloom} for multiple choice tasks. As both approaches aim to expand context for ICL, we incorporate more examples (\(W=18\)) to evaluate their ability to handle extensive context. From Tables \ref{tab:nli_bloom-7b1} and \ref{tab:multi_choice_bloom}, several observations emerge: 
(1) Unlike PCW, both StructedICL and our MateICL approaches benefit from expanded context, enhancing accuracy and stability. In the worst-case scenario, their performance remains stable, demonstrating the idea of revisiting attention weights of task tokens.
(2) MateICL generally outperforms StructedICL in terms of accuracy and stability and the improvements can be deemed significant.
(3) We notice a decrease in StructedICL's performance as context greatly increases (\(W=18\)). This decline is expected, as StructedICL simply sums up attention weights of task tokens by a factor of \(W\), causing excessive self-focus and neglect of context.
\begin{figure}
	\centering
	\includegraphics[scale=0.29]{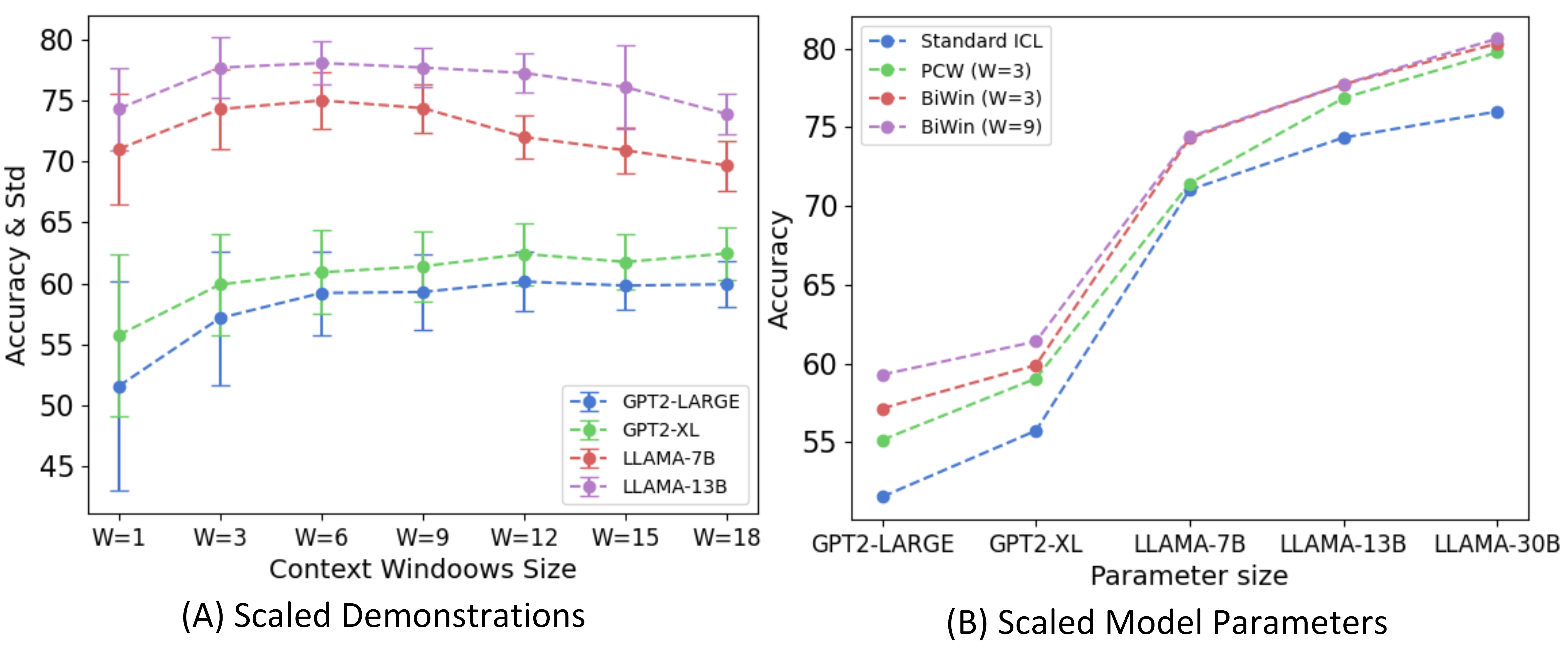}
	
	\caption{(A) Averaged accuracy and error bars for scaled demonstrations and (B) model size correlation with average text classification improvements.}

	\label{fig:correl_model_impr_text}
\end{figure}

\subsection{Sensitivity Evaluation}
In this section, we examine the sensitivity of MateICL to various parameters.

\subsubsection{Impact of Varied \(b\) of Eq.\ref{eq:bias_value}}\label{sec:biases}
We explored various \(b\) using text classification tasks across all models, maintaining a constant context size, to evaluate their impact. Fig.\ref{fig:various_bias} shows average performance (A) and stability (B)   for \(W=\{3,6,9\}\). Results reveal that (\(b=1\)) leads to the poorest performance. Optimal performance is achieved with moderate \(b\) (\(3\leq b\leq 5\)), validating our \(b\) choice. Furthermore, we notice a trade-off: increased \(b\) slightly improves stability but may slightly reduce accuracy.

\subsubsection{Effect of Scaled Demonstrations}
Fig.\ref{fig:correl_model_impr_text}.A shows that GPT2, a smaller model, is less affected by larger contexts (\(1\leq W\leq 18\)), processing 684 demonstrations at (\(W=18\)) in TREC, compared to LLaMA's 1,242. Fig.\ref{fig:correl_model_impr_text}.B shows that increasing demonstrations (e.g., \(W=9\)) improves model stability in ICL, aligning with \cite{ZhaoWFK021}.

\subsubsection{Effect of Model Size}
Fig.\ref{fig:correl_model_impr_text}.B shows the correlation between model size and text classification performance improvements.  Larger models yield significant enhancements, However, these improvements come with an increase in computational cost.

\section*{Limitations}

While MateICL enhances ICL by enabling more demonstrations without requiring fine-tuning, it has notable constraints:

\begin{itemize}
	\item In tasks demanding sequential or interrelated contexts, such as in code generation, its effectiveness is limited. This is due to MateICL's inherent structure, which may struggle with handling dependent or sequential information.
	\item Merely increasing the number of shots does not guarantee accuracy improvements across all scenarios. Studies show that enlarging the context window size does not consistently yield better results in completion tasks.
\end{itemize}

\section{Conclusion}\label{sec:conclusion}

This paper introduces Mitigating Attention Dispersion in large-scale ICL (MateICL) framework, which enables models to maintain effective self-attention as the number of demonstrations increases.  Our experiments showed MateICL's superiority over baselines. MateICL also benefits from supervision, enhancing stability.  For future work, MateICL could further enhance ICL with more examples in fine-tuning setups;  However, further investigation is needed.

\bibliography{anthology}

\appendix
\section{Experimental Setup}\label{sec:setup}
\subsection{HuggingFace Attention Modification} \label{sec:code}
We update the multi-head attention as follows:\\
{\footnotesize
\begin{lstlisting}[language=Python]
if W > 1:   
	past_key_values_length = key_states.shape[-2]       
	v = (int((W / 3)) + 2) if W > 3 else 2    
	b = torch.ones(1, 1, 1, key_states.shape[-2])
        b = b.to(key_states.device)
	b[:, :, :, past_key_values_length:] = v
	attn_weights = attn_weights * b
	attn_weights/=attn_weights.sum(dim=-1, keepdim=True)
\end{lstlisting}}

\section{Experimental Setup}\label{sec:appendix_experiment}
\begin{table}[t]
	\centering
	
	\adjustbox{width=\linewidth}{		
		\begin{tabular}{l|l|l}
			\hline
			Dataset & \makecell[c]{Prompt Example} & Labels \\
			\hline	
			CB 		&\makecell[l]{Premise:\{Premise\}\textbackslash n\\Hypothesis:\{hypothesis\}\textbackslash n\\ Entailment:\{Label\}}&[true, false, neither]\\
			\hline 		
			WiC		&\makecell[l]{Statement 1:\{sentence1\}\textbackslash n\\Statement 2:\{sentence1\}\textbackslash n\\Answer:\{Label\}}&[yes, no]\\			
			\hline 
			BoolQ	&\makecell[l]{Premise:\{Passage\}\textbackslash n\\Hypothesis:\{Question\}\textbackslash n\\ Answer:\{Label\}}&[yes, no]\\			
			\hline 
			RTE		&\makecell[l]{Premise:\{premise\}\textbackslash n\\Hypothesis:\{hypothesis\}\textbackslash n\\ Entailment:\{Label\}}&[yes, no]\\			
			\hline 
			SCITAIL		&\makecell[l]{Statement 1:\{sentence1\}\textbackslash n\\Statement 2:\{sentence1\}\textbackslash n\\Answer:\{Label\}}&[yes, no]\\			
			\hline 
			QNLI	&\makecell[l]{Sentence:\{sentence\}\textbackslash n\\Question:\{question\}\textbackslash n\\ Answer:\{Label\}}&[yes, no]\\			
			\hline 
			SICK	&\makecell[l]{Statement 1:\{sentence\_A\}\\Statement 2:\{sentence\_B\}\textbackslash n\\ Entailment:\{Label\}}&[true, false, neither]\\			
			\hline 
			SNLI	&\makecell[l]{Premise:\{Sentence\}\textbackslash n\\Hypothesis:\{hypothesis\}\textbackslash n\\ Entailment:\{Label\}}&[true, false, neither]\\			
			\hline
		\end{tabular}
	}
	\caption{The prompt format of Natural Language Inference datasets.}
	\label{tab:format_nli}
\end{table}
\begin{table}[t]
		\centering
		
			\begin{tabular}{l|l}
				\hline
				Dataset & \makecell[c]{Prompt Example} \\
				\hline	
				SQuAD 		&\makecell[l]{Title:\{title\}\textbackslash n\\\{context\}\textbackslash n\\Question:\{question\}\textbackslash n\\Answer: \{answers\}}\\
				\hline 		
				SQuADV2 		&\makecell[l]{Title:\{title\}\textbackslash n\\\{context\}\textbackslash n\\Question:\{question\}\textbackslash n\\Answer: \{answers\}}\\
				\hline 		
				WebQS 		&\makecell[l]{Question:\{question\}\textbackslash n\\Answer: \{answers\}}\\
				\hline 							
			\end{tabular}
		
		\caption{The prompt format of MRC datasets.}
		\label{tab:format_mrc}
	\end{table}

\begin{table}
	\centering
	
	\adjustbox{width=\linewidth}{		
		\begin{tabular}{l|l|l}
			\hline
			Dataset &\makecell[c]{Prompt Example} & \#Choices \\
			\hline	
			PIQA 		&\makecell[l]{\{goal\}\textbackslash n\\\{label\}}&4\\
			\hline 		
			OBQA 		&\makecell[l]{\{question\_stem\}\textbackslash n\\\{answerKey\}}&4\\
			\hline 		
			COPA 		&\makecell[l]{\{goal\}\textbackslash n\\\{label\}}&4\\
			\hline 		
				HeLLaSwag 	&\makecell[l]{\{activity\_label\}\textbackslash n\\\{ctx\}\textbackslash n\\\{endings\}\textbackslash n\\\{label\}}&4\\
				\hline 		
				ARCE 		&\makecell[l]{\{question\}\textbackslash n\\\{answerKey\}}&4\\
				\hline 		
				StoryCloze 	&\makecell[l]{\{InputSentence1\}\textbackslash n\\\{InputSentence2\}\textbackslash n\\\{InputSentence3\}\textbackslash n\\\{InputSentence4\}\textbackslash n\\\{AnswerRightEnding\}}&4\\
				\hline 		
				MMLU 		&\makecell[l]{\{question\}\\\{answer\}}&4\\
				\hline

			\end{tabular}
		}
		
		\caption{The prompt format of Multiple Choices Question Answering datasets.}
		\label{tab:format_multi}
	\end{table}
	
	\begin{table}[t]
		\centering
		
			\begin{tabular}{l|l}
				\hline
				Dataset & Prompt Example  \\
				\hline	
				SST-2 &  \makecell[l]{Sentence: \{Sentence\}\\Label: {Label}}  \\
				\hline
				
				CR	&\makecell[l]{Review:\{Sentence\}\\ Sentiment:\{Label\}}\\
				\hline 		
				SUBJ	&\makecell[l]{Input:\{Sentence\}\\ Type:\{Label\}}\\
				\hline 			
				CB 		&\makecell[l]{Premise:\{Sentence\}\\Hypothesis:\{\ hypothesis\}\\ Prediction:\{Label\}}\\
				\hline 		
				RTE	&\makecell[l]{Premise:\{Sentence\}\\Hypothesis:\{\ hypothesis\}\\ Prediction:\{Label\}}\\
				\hline 	
				AGNews	&\makecell[l]{Input:\{Sentence\}\\ Type:\{Label\}}\\
				\hline 		
				SST-5&\makecell[l]{Review:\{Sentence\}\\ Sentiment:{{{Sentiment}}}}\\
				\hline 	
				TREC &\makecell[l]{Question:\{Sentence\}\\ Type:\{Label\}}\\
				\hline 							
				DBPedia		&\makecell[l]{Input:\{Sentence\}\\ Type:\{Label\}}\\
				\hline 	
			\end{tabular}
		
		\caption{The prompt format of Text Classification datasets.}
		\label{tab:format_class}
	\end{table}

\subsection{Prompt Format} \label{sec:prompt}
We present the prompt formats for datasets in Text Classification, NLI, Multiple-Choice Question Answering, and Machine MRC in Tables  \ref{tab:format_nli}, \ref{tab:format_mrc}, \ref{tab:format_multi}, and \ref{tab:format_class}, respectively. It is important to mention that we have not adopted any form of instruction engineering in the development of these prompt formats. Some of these formats were adapted from GPT-3 \cite{BrownMRSKDNSSAA20} and PCW \cite{pcw}. Note that we adopt the same templates of \citet{Yaru-2212-06713} when experimenting on Bloom.

\end{document}